\documentclass[review]{elsarticle}

\usepackage{color}
\usepackage{amsmath}
\usepackage{amssymb}
\usepackage{latexsym}
\usepackage{array}
\usepackage{multirow}
\usepackage[table]{xcolor}
\usepackage{hhline}

\newcolumntype{L}[1]{>{\raggedright\let\newline\\\arraybackslash\hspace{0pt}}m{#1}}
\newcolumntype{C}[1]{>{\centering\let\newline\\\arraybackslash\hspace{0pt}}m{#1}}
\newcolumntype{R}[1]{>{\raggedleft\let\newline\\\arraybackslash\hspace{0pt}}m{#1}}

\DeclareMathOperator*{\argmin}{arg\,min}

\newcommand{\gbc}[1]{{\cellcolor{green!25}\textbf{#1}}}

\usepackage{lineno,hyperref}
\modulolinenumbers[5]

\begin{document}

\begin{frontmatter}

%\title{Elsevier \LaTeX\ template\tnoteref{mytitlenote}}
\title{AFIF$^4$: Deep Gender Classification based on AdaBoost-based Fusion of Isolated Facial Features and Foggy Faces}
%\tnotetext[mytitlenote]{Fully documented templates are available in the elsarticle package on \href{http://www.ctan.org/tex-archive/macros/latex/contrib/elsarticle}{CTAN}.}

%% Group authors per affiliation:
\author[mymainaddress,mysecondaryaddress]{Mahmoud Afifi}%\fnref{myfootnote1}
%\address{EECS, Lassonde School of Engineering, York University}
%\fntext[myfootnote1]{Corresponding author}
\ead{mafifi@eecs.yorku.com}
\cortext[mycorrespondingauthor]{Corresponding author}

\author[mymainaddress,mysecondaryaddress]{Abdelrahman Abdelhamed\corref{mycorrespondingauthor}}
%\address{EECS, Lassonde School of Engineering, York University}
%\fntext[myfootnote2]{}
\ead{kamel@eecs.yorku.com}
\cortext[mycorrespondingauthor]{Corresponding author}

%% or include affiliations in footnotes:
%\author[mymainaddress,mysecondaryaddress]{Elsevier Inc}
%\ead[url]{www.elsevier.com}

%\author[mysecondaryaddress]{Global Customer Service\corref{mycorrespondingauthor}}
%\cortext[mycorrespondingauthor]{Corresponding author}
%\ead{support@elsevier.com}

\address[mymainaddress]{Department of Electrical Engineering and Computer Science, Lassonde School of Engineering, York University, Canada}
\address[mysecondaryaddress]{Faculty of Computers and Information, Assiut University, Egypt}

\begin{abstract}
Gender classification aims at recognizing a person's gender. Despite the high accuracy achieved by state-of-the-art methods for this task, there is still room for improvement in generalized and unrestricted datasets. In this paper, we advocate a new strategy inspired by the behavior of humans in gender recognition.      Instead of dealing with the face image as a sole feature, we rely on the combination of isolated facial features and a holistic feature which we call the \textit{foggy face}. Then, we use these features to train deep convolutional neural networks followed by an AdaBoost-based score fusion to infer the final gender class. We evaluate our method on four challenging datasets to demonstrate its efficacy in achieving better or on-par accuracy with state-of-the-art methods. In addition, we present a new face dataset that intensifies the challenges of occluded faces and illumination changes, which we believe to be a much-needed resource for gender classification research.
\end{abstract}
\end{frontmatter}

%\linenumbers

\section{Introduction}

Gender classification is undoubtedly a simple task for humans, however, it is still an active research problem that draws the attention of many researchers in various fields, including computer vision and machine learning, with many applications \cite{ng2015review,jain2011handbook}, such as visual surveillance, intelligent human-computer interaction, social media, demographic studies, and augmented reality.     

Although human faces are a powerful visual biometric feature, some facial features have semantic structures that may mislead the classification process relying on facial images as we show later in Section \ref{sec:method}.  In this paper, we advocate a different strategy to address the gender classification problem by mimicking the human behavior in gender recognition.    We started by conducting some user studies to obtain a good grasp on how facial features can sometimes be unreliable in gender recognition.   Then, we show how the human behavior in gender recognition can help us decide which facial features are more reliable.    To this end, instead of dealing with the raw face images, we extract a few reliable facial features, then we input each of these features to a deep convolutional neural network (CNN) that initially classifies each visual feature to be belonging to either male or female.   Finally, we use an AdaBoost-based score fusion to get the final classification decision based on the prediction score of each separate facial feature (Section \ref{sec:method}).     To evaluate the efficacy of our method, we apply it to four widely-used gender classification datasets, what reveals that our method achieves better, or at least on-par, results than many state-of-the-art methods (Section \ref{sec:results}).

With the state-of-the-art gender classification methods achieving compelling accuracy on the existing benchmark datasets, we believe that there is a need for more datasets focusing on more challenging scenarios for gender classification.   So, in addition to the proposed method, we also propose a new challenging face dataset where we focus mainly on challenging cases such as occluded  and badly illuminated faces.    We also evaluate our method on this proposed dataset, revealing how it can be more challenging than other existing datasets.   The rest of the paper is as follows: in Section \ref{sec:related-work}, we provide a quick review on key gender classification methods and discuss some of the common shortcomings. In Section \ref{sec:method}, we discuss in depth our proposed method. Then, we present our proposed face image dataset and experimental evaluation of our method in Section \ref{sec:results}, followed by a brief conclusion in Section \ref{sec:conclusion}.

\section{Related Work} \label{sec:related-work}

Vision-based gender classification methods are usually based on extracting features from the given face image then use these features to train a classifier that outputs the predicted gender. Such methods, can be divided into two main categories: 1) geometric-based, and 2) appearance-based methods.

The geometric-based techniques extract and utilize geometric features from the given face image to predict the gender. Burton et al. \cite{burton1993} presented a gender classification technique that relies on 73 facial points and uses discriminant analysis of point-to-point distances to infer the gender. Hands were used as biometric traits by Amayeh et al. \cite{amayeh2008} where the extracted geometric features of different parts of the hand were used for gender discrimination. The main issue with such methods is that they require highly precise extraction of the geometric features to obtain good classification accuracy.

On the other hand, appearance-based methods rely on extracting features from either or both of: i) the whole face image (holistic features) and ii) regions of the face image (local features). Li et al. \cite{gender} introduced a method based on five individual facial features in addition to the hair and clothing of the person then used multiple support vector machine (SVM) classifiers to classify the gender based on the individual features. By combining this elaborated visual information using different five fusion approaches, they improved the classification accuracy even more. Nevertheless, the clothing and hair information may mislead the classifier, since the lack of context-based representation can be tricky even for humans. 

Combining both geometric-based and appearance-based features, Mozaffari et al. \cite{mozaffari2010} presented a technique that extracts both geometric-based and appearance-based features using three alternative methods: discrete cosine transform, local binary pattern (LBP), and geometrical distance.  Tapia et al. \cite{tapia2013} proposed a technique that uses fusion of local visual features where the feature selection is based on the information theory. They used the mutual information to measure the similarity of pixel intensities in order to reduce redundant features. They applied this measure on three different features: pixel intensities, shape, and local image texture that was described using LBPs. 
 
Rai et al. \cite{rai2014} presented a gender classification system that uses local visual information extracted from the Region of Interest (ROI) which is determined manually using three selected points. The ROI is divided into grid sub-images that were used to generate Gabor space by applying 2D Gabor filter with six orientations on each sub-image in order to reduce the sensitivity to light variations. The reduced feature vector is used as an input to SVM classifier that distinguishes between male and female local features. 

Hadid et al. \cite{hadid2015} showed the efficiency of using LBPs in the gender and texture classification. Recently, Castrill{\'o}n-Santana et al. \cite{castrill2016} presented a comparative study among ten local feature (holistic and local) descriptors and three fusion strategies for gender classification using two datasets, namely EGA \cite{ega} and Groups \cite{groups}. They reported that local salient patterns (LSP) \cite{lsp} and histogram of oriented gradients (HOG) \cite{hog} achieve the best accuracy using holistic visual features, while local phase quantization (LPQ) \cite{lpq} with SVM attains the best accuracy using local visual features. Moeini and Mozaffari \cite{moeini2017} proposed to learn separate dictionaries of male and female features, extracting  64 feature vectors of the face image using LBP. Then, a sparse representation-based classifier is used in the classification process, reporting state-of-the-art accuracy.

On the other end, deep neural networks are becoming increasingly ubiquitous in many classification problems, according to the achieved remarkable improvements on accuracy. Levi and Hassner \cite{levi2015} classified both gender and age via a simple Convolutional Neural Network (CNN) that was applied to the Adience benchmark \cite{Adience} for age and gender classification in a holistic manner. A local CNN was used by Mansanet et al. \cite{mansanet2016} where they utilized Sobel filter in order to extract the local patches while taking into account the location of each patch. In order that the trained CNN obtains high accuracy with face images under occlusions, Juefei-Xu et al. \cite{juefeixu2016} utilized multiple levels of blurring to train a deep CNN in a progressive way. 

From the above quick review, we can see that most of the prior work in gender classification was depending on extracting large number of local features per image, hand-crafted features, or unreliable holistic features. On the contrary, in our approach, we avoid such issues by using only four highly-discriminative local features and one holistic feature, and we combine these features with the classification power of deep CNNs to achieve state-of-the-art, or at least on-par, gender classification accuracy. As we discussed in the introduction, our choice of facial features is mainly based on the findings from the user studies we carried out trying to understand how humans behave in the gender recognition process. A detailed description of these user studies and our full approach follows in Section \ref{sec:method}.

\section{Our Methodology} \label{sec:method}

In this section, we will discuss in detail our approach to addressing the gender classification problem. First, we discuss the user studies carried out to help us decide which facial features (local and holistic) are more discriminative for gender classification. Then, we show how we prepare the face patches using a deformable part-based model as proposed by Yu et al. \cite{cdsm}. After that, we discuss the training of CNNs to classify the facial patches and get initial classification decisions.  Finally, we present an adapted AdaBoost-based fusion mechanism of initial classification labels leading to the final classification decision.

\subsection{A quick study of human behavior in gender recognition}

To study how the gender discrimination is performed by humans, we carried out two experiments. For these experiments, we used 200 images from the Labeled Faces in the Wild (LFW) dataset \cite{lfw}, containing both male and female faces. The ground truth classes (male or female) was based on the attribute classifiers presented by Kumar et al. \cite{LFWAttributes}.

% were carried out using 200 images, for both males and females. The images were picked from the Labeled Faces in the Wild (LFW) dataset (\cite{lfw}) based on the attribute classifiers presented by \cite{LFWAttributes}. All images have a lower gender attribute, for example male faces of high attractiveness (see Subsection \ref{LFW_subsection}). Thus, it is not a simple task to discriminate the gender of individuals in the picked images. 

In the first experiment, we wanted to see how a human decides on the gender of a face image, especially when the facial features are not very discriminative. To do that, we showed 100 images to 5 different volunteers (20 images per volunteer) in two stages. In the first stage, each volunteer was asked to watch 10 different images and classify each one as male or female.   Before doing the second stage, the volunteers were asked to be ready to explain what they would do once they are uncertain about the gender of the face. At the end of the experiment, almost all of the volunteers reported that the first thing they did was to look at the face region and facial features (eyes, nose, ears, etc.), if they cannot precisely determine the gender, they look at the whole image and think about all of the visual information (clothing, hair, accessories, etc.) surrounding the face in order to make their final decision. It is worth noting that the classification accuracy in this experiment was 96\%.

\begin{figure*}[t!]
\centering
\includegraphics[width=1\linewidth]{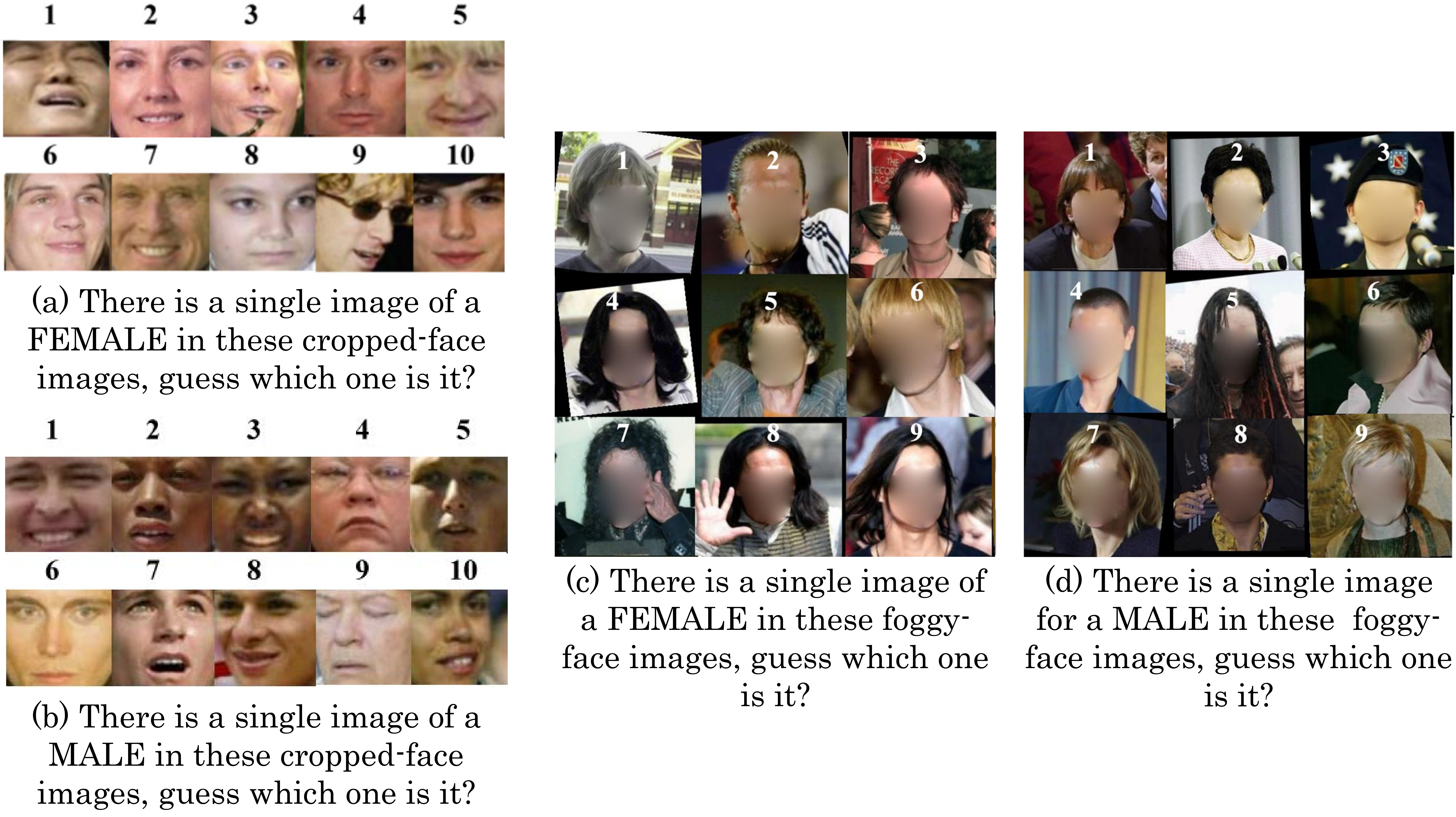}
\caption{\label{fig:user-study}Sample questions from our user studies. (a) and (b) are asking the users to guess a male/female image among a set of female/male images using cropped-face images. (c) and (d) are the same questions using foggy-face images. All images are taken from the Labeled Faces in the Wild (LFW) dataset \cite{lfw}.}
\end{figure*}

From the first experiment, we notice that visual information surrounding the face in an image are of high importance in classifying the gender of the face, especially when the facial features are quite ambiguous. That led us to conduct the second experiment, where we wanted to see which is more discriminative     in deciding the gender from a human perspective: the facial features or the visual information surrounding the face? To achieve this, we used two types of images: 1) \textit{cropped face} images that contain only the facial features, and 2) \textit{foggy face} images that contain the whole visual information surrounding the face while the face region being heavily blurred out, so that a volunteer will depend only on the surrounding visual information to decide on the gender. Figure \ref{fig:user-study} shows some examples of the two types of images.       Then, we prepared an on-line subjective test that asks volunteers to guess the image of a male among a set of female images and vice versa, i.e., to guess the image of a female among a set of male images. Each question contained either 5 or 10 images from either the cropped or the foggy face images.  There were 70 volunteers who accomplished this experiment using 100 images.  Additionally, the volunteer is asked to comment on which type of face images was easier to recognize its gender.        As a result, we got classification accuracy of 69.40\% for the cropped face images and a higher accuracy of 88.09\% for the foggy face images. Also, most of the volunteers reported that the foggy face images were easier to classify than the cropped face images.

%Note that, all images were hard to classify and have many misleading features that are surprisingly tricky. Besides, the images where either contain only the cropped face region or a blurred face image (without facial features). 

%hair and isolated facial features have a great impact on the classification process (\cite{hair}; \cite{face}).

From the above experiments, we notice that, beside the high importance of isolated facial features, the visual information from the general look of persons also possesses an important role in the classification process regardless of the visibility of facial features.  That was also noticed in prior work by Lian and Lu \citep{hair}. As a result, we decided to build our approach to gender classification by combining both isolated facial features and general appearance of the subject in the classification process, as we will discuss in Section \ref{subsec:preparation}.

%Hence, a conclusion can be driven that the human visual system precepts all visual information and the brain classifies the gender based on the combination of the isolated visual information with taking into account the general appearance of the subject. 

\subsection{Facial features preparation}    \label{subsec:preparation}

Before extracting the facial features and to improve the face detection process involved in our approach, we preprocess the images by applying lighting-invariant enhancement techniques. This was inspired from the work by Han et al. \cite{preprocessingFaces}. We apply the single scale retinex (SSR) presented by Jobson et al. \cite{SSR}, in which the lighting-invariant enhanced image $I_{SSR}$ (the retinex image) is generated by subtracting the estimated global illumination from the given face image $I$ in the log space, as in the following equation:
\begin{equation}
I_{SSR}(x,y)= \log (I(x,y)) - \log(F(x,y) \ast I(x,y)),
\end{equation}%
where $(x,y)$ represents the spatial location of a pixel in the image, $F$ is the Gaussian low-pass filter given by%
\begin{equation}
F(x,y) = K \, e^{-(x^2+y^2)/G^2},
\end{equation}%
in which $K$ is determined such that $\iint F(x,y) \, dx \, dy = 1$ and $G$ is the Gaussian surround contrast.

\begin{figure*}[t!]
\centering
\includegraphics[width=1\linewidth]{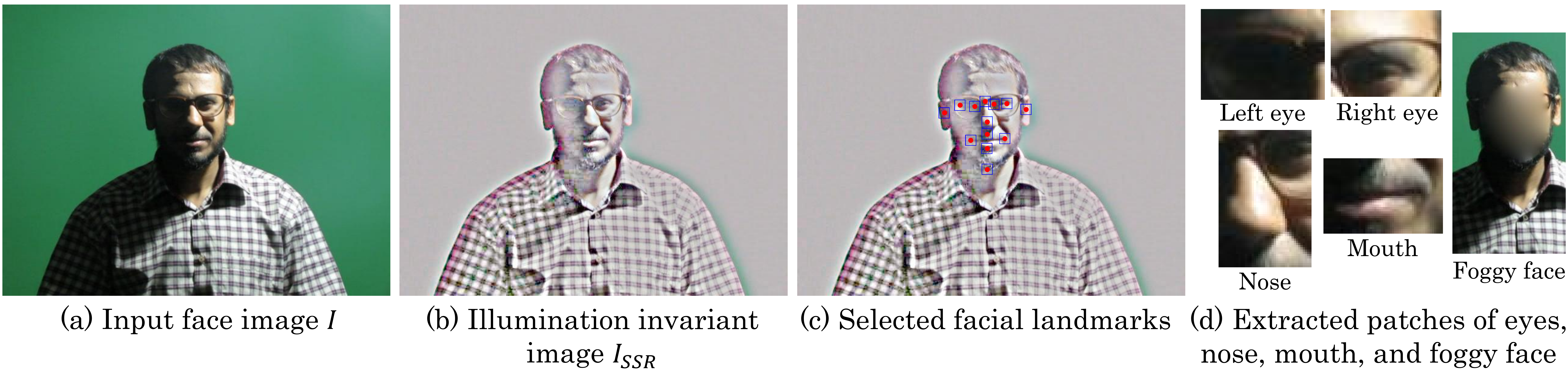}
\caption{\label{fig:feature-extraction}The process of extracting facial features including the foggy face. (a) is the input image. (b) is the illumination-invariant image generated by the single scale retinex (SSR) method \cite{SSR}. (c) is the selected facial landmarks from the face detection process by fitting the cascaded deformable shape model (CDSM) \cite{cdsm}. (d) is the extracted features: eyes, nose, mouth, and foggy face which is generated by applying the Poisson image editing (PIE) technique \cite{poisson}.}
\end{figure*}

In the next step, we scan the input image $I$ to fit the cascaded deformable shape model (CDSM) proposed by Yu et al. \cite{cdsm} in order to detect and extract the face regions. Fitting a CDSM requires the maximization of a scoring function that is based on both local appearance and shape optimization. This score function represents how well the estimated facial landmark positions are aligned with the CDSM.  In order to detect multiple faces in a single image (as in the case of the Groups dataset), we repeat the following 2-step procedure: 1) we apply the CDSM to detect a face; 2) we mask out the detected face region by a single-color rectangle and repeat step 1 again on the same image until no more faces are detected.  In the case that applying the CDSM on the original image $I$ fails, we use the SSR image $I_{SSR}$ instead. Finally, we use the estimated landmarks of eyes, nose, and mouth, to extract a separate patch for each visual feature. Figure \ref{fig:feature-extraction} shows an example of the process of extracting facial patches.

% extra areas, whose thickness equal to 30\% of patch's dimensions, are added to avoid the mean average pixel error of the facial landmarks. 

To generate the foggy face images, first, the face region is detected using the above mentioned procedure, then the surrounding facial landmarks are used to define an unknown region $\Omega$ which we feed into a Poisson image editing (PIE) equation \cite{poisson}:
\begin{equation}
\label{possionEq}
\argmin_f \iint_{\Omega}^{} \lvert \nabla f \rvert ^2 \quad \text{with} \quad f|_{\partial \Omega}=f^\ast|_{\partial \Omega},
\end{equation} 
where $\nabla f$ is the first derivative of the unknown scalar function over $\Omega$ in the given image $I$. By omitting the suggested guidance used by \cite{poisson}, the foggy face image is generated by solving Equation \ref{possionEq}. Figure \ref{fig:feature-extraction}d shows an example of a generated foggy face image. 

\subsection{Facial feature classification using CNNs}

Once we have all the facial features ready, we feed each patch as an independent input to a pre-trained CNN that is dedicated to classifying this biometric trait. In other words, we train four separate CNNs for the four separate facial features: foggy face, eyes, nose, and mouth. We adapted the Caffe reference model, proposed by Jia et al. \cite{caffe}, to solve the binary classification problem of each visual patch. Our adapted CNN architecture consists of five convolutional layers and three fully-connected layers. The first convolutional layer contains 96 (11$\times$11) convolutional filters and uses stride size equal to 4 to reduce the computational complexity. The second layer uses one stride with 256 (5$\times$5) convolutional filters. The last three convolutional layers use one stride with 9 (3$\times$3) filters. The final softmax layer responds with two possible output classes, representing either male or female. The CNN is trained using stochastic gradient descent and back-propagation \cite{back} over 1000 iterations. %\red{Our adapted CNN architecture is illustrated in in Figure \ref{XX}}.

\subsection{Classification score fusion}

\begin{figure*}[t!]
\centering
\includegraphics[width=1\linewidth]{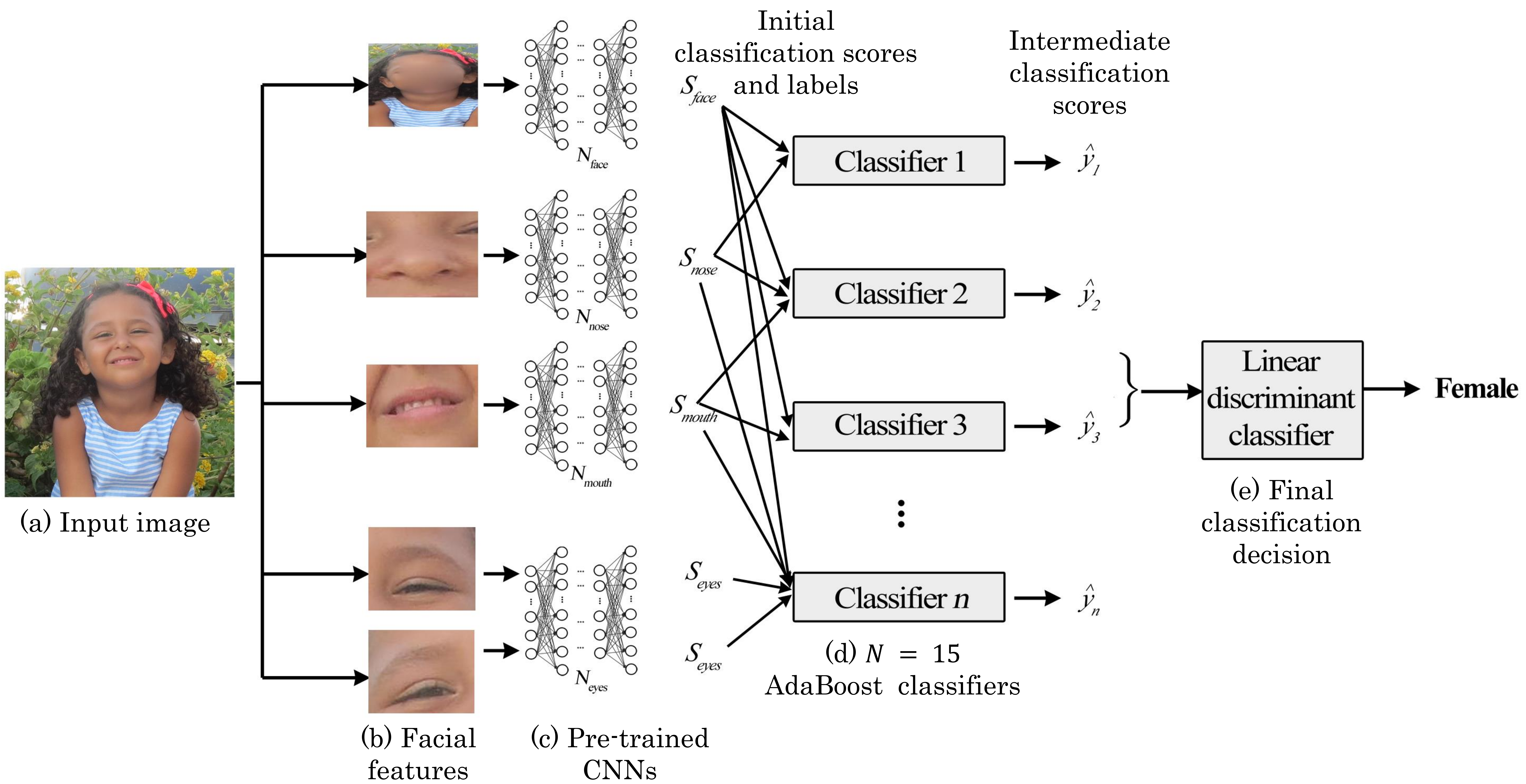}
\caption{\label{fig:overview}An overview of our whole approach for gender classification. (a) the input image. (b) the extracted facial features (foggy face, eyes, nose, and mouth patches). (c) initial classification scores from 4 pre-trained convolutional neural networks (CNNs). (d) intermediate classification decisions based on 15 weak classifiers, one classifier for each combination from the initial scores. (e) the final classification decision from a linear discriminant classifier.}
\end{figure*}

Up to now, we have four separate CNNs that give us four independent gender classification scores based on four separate facial features. To fuse the four independent classification scores, we apply an AdaBoost-based score fusion mechanism as follows.    Let $N_L$ be one of the five pre-trained CNNs that gives a decision $c$, such that $c \in \{\text{MALE}=1, \text{FEMALE}=-1\}$, and $s_L$ is the corresponding prediction score given by the softmax function of $N_L$, where $L \in \{\text{face}, \text{eyes}, \text{nose}, \text{mouth}\}$ and $S_L=c(s_L$).  We use the foggy face score $S_{face}$ that represents the prediction score of the foggy face multiplied by the estimated class as a holistic score that is combined with all other possible feature scores. Since we have four more scores (for left eye, right eye, nose, and mouth), we get 15 different combinations of scores, which is the summation given by
 \begin{equation}
 \label{eq:comb}
\sum_j \binom{| L |}{j}, \quad j \in \{ 1, \dots, |L|\},
\end{equation}%
where $|L|$ is the cardinality of the set of features. Then, we train AdaBoost classifiers \cite{ada} using the combination vectors to get the estimated class $\hat{y_i}$ of each combined $i^{th}$ vector, where $i \in \{1,\dots,15\}$. For each vector $\vec{v_i}=<S_{\text{face}},\dots>$, the predicted class $\hat{y_i}$ is given by
 \begin{equation}
 \label{adaeq}
\hat{y_i}= \text{sign} (\sum_{t=1}^{T}\alpha_{t}C_{t}(\vec{v_i})),
 \end{equation}
where $C_t(\vec{v_i})$ is the estimated class of the given score vector $v_i$ using the $t^{th}$ weak classifier, $\alpha_{t}$ is the output weight of the classifier, and $T$ is the number of weak classifiers. Eventually, the suggested labels are combined into a single vector $\vec{\hat{Y}}=<\hat{y_1},\hat{y_2},\dots,\hat{y_n}>$ to train a linear discriminant classifier that determines the final class of the face image.   An overview of our whole approach is illustrated in Figure \ref{fig:overview}.

In this section, we have discussed our approach to the gender classification problem, from deciding on the features we used to the final classification decision. In section \ref{sec:results}, we present some experimental results to evaluate our approach and compare it against other state-of-the-art methods, showing its high performance and efficacy.

\section{Experimental Results} \label{sec:results}

In this section, we demonstrate the efficacy of our approach through extensive evaluation against four widely-used benchmark datasets. Additionally, we evaluate our approach against our new challenging dataset, the \textit{Specs of Faces} dataset.

\subsection{Evaluation Setup}
In our evaluation procedure, we use five-fold cross validation on all datasets and report the mean of the accuracy values. As the suggested folds of some datasets contain unbalanced number of male and female images, random images are picked from the excessive group in order to have the same number of images for both genders. We train the deep CNNs using feature patches of size $227\times 227$ pixels, extracted from 75\% of the training set. The AdaBoost classifiers are trained thereafter by 60\% of the rest of the training set using the prediction scores of the pre-trained CNNs. At the final stage of training, the fusion classifier is trained using the estimated classes reported by the AdaBoost classifiers over the rest of the training set. Eventually, we test the entire algorithm using the testing fold.

\begin{figure*}[t!]
\centering
\includegraphics[width=.65\linewidth]{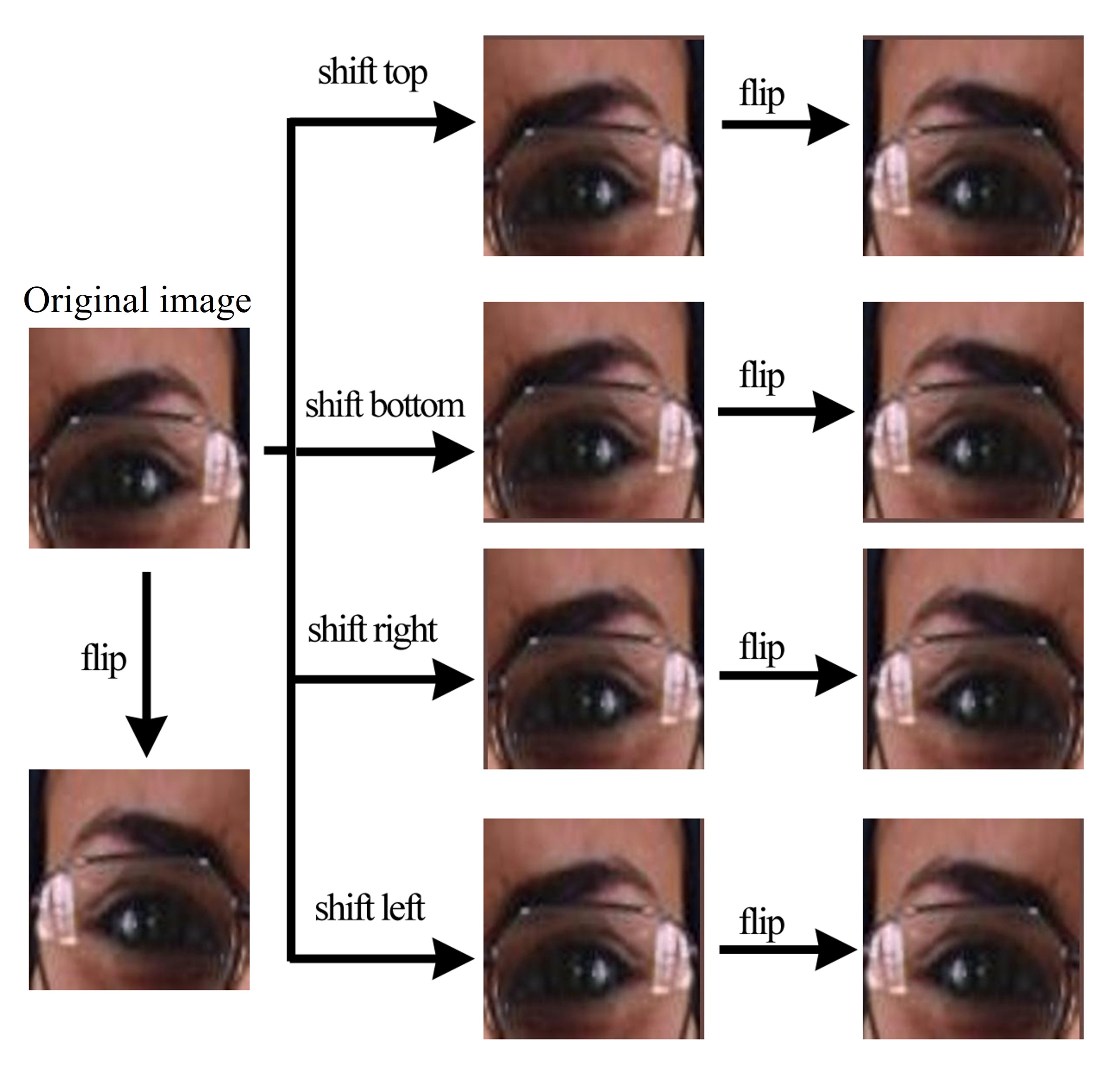}
\caption{\label{fig:enlargeTrainingImage}An example on how we enlarge the number of training images by applying small translation on the original image along the four sides, then horizontally flip the four translated images and the original image. The shown image is an eye patch from the proposed Specs on Faces dataset.}
\end{figure*}

As the CNNs are usually susceptible to the overfitting problem using a limited number of images, we enlarged the number of training images 10 times by generating more images through applying a set of operations depicted in Figure \ref{fig:enlargeTrainingImage}. For each training image, we apply translation by 5 pixels along the four border sides, then horizontally flip each of the four translated images and the original image. The empty pixels that have been produced by the previous operations are filled by the mean of the original training image to maintain the equilibrium of original means over the training set.

\subsection{Datasets}
\label{datasets}
We have evaluated our method against four challenging datasets: the Labeled Faces in the Wild (LFW) \cite{lfw}, the Images of Groups dataset \cite{groups}, the Adience benchmark for age and gender classification \cite{Adience}, and the Face Recognition Technology dataset (FERET) \cite{FERET}. Furthermore, we present a new dataset of challenging face images and use it to evaluate our method. The proposed dataset is denoted as the Specs on Faces (SoF) dataset. 

\noindent\textbf{Labeled Faces in the Wild (LFW).} The LFW dataset \cite{lfw} consists of 13,233 unconstrained face (250$\times$250 pixels) images for different 5,749 persons (4,272 males and 1,477 females). In order to label the images based on gender, we used the attribute values presented by Kumar et al. \cite{LFWAttributes}. Each descriptive visual attribute $Z$ is represented as a real value $a_z: a_z\in \mathbb{R}$, where the magnitude of $a_z$ represents the degree of $Z$ and the sign of $a_z$ represents the category. In the gender attribute, a positive sign refers to a male image and a negative sign represents a female image. It is worth noting that there is a reported error rate (approximately 8.62\%) in this classification. A straightforward way was used to assign each face image to its gender label by applying a threshold based on the sign of the gender attribute. However, there are some images whose gender attributes lie on the boundaries (e.g. $\pm$0.3); that leads to incorrect labels. To handle that, we added another layer of separation for images whose magnitude values are less than a threshold (e.g. 0.5). Then, we used the \texttt{genderize.io}\footnote{\url{https://genderize.io/}} API to estimate the gender based on the first name of each face image in the LFW. Eventually, we manually reviewed each category of male and female images three times to completely eliminate any incorrect labels. We made this accurate labeling of the LFW dataset available online\footnote{\url{http://bit.ly/lfw-gender}}. In our experiments, we used 2,948 images from the LFW dataset (590 images on average for each fold).

\noindent\textbf{Images of Groups dataset.} The Groups dataset \cite{groups} contains 28,231 face images that were extracted from the original 5,080 group images collected from Flickr images. The Groups dataset is considered, in the literature, the most challenging and complex dataset for the gender classification problem \cite{onsoft,FRVT,ref10}. The experiments were carried out using 12,682 face images (2,536 images on average for each fold).

\noindent\textbf{Adience benchmark for age and gender classification.} The Adience benchmark \cite{Adience} comprises 26,580 unconstrained face images gathered from Flickr albums for 2,284 persons. The images include people with different head poses and ages under various illumination conditions. The folds which have been used in the experiments were picked randomly, where each fold contains 970 images on average.

\noindent\textbf{Face Recognition Technology (FERET).} The FERET dataset \cite{FERET} is widely used to evaluate and develop facial recognition techniques. The dataset consists of 14,126 images for 1199 different persons captured between 1993 and 1996. There is a variety in face poses, facial expressions, and lighting conditions. In 2003, the high resolution (512$\times$ 768 pixels) color FERET was released which has been used in the presented experiments. The total number of frontal and near-frontal face images, whose pose angle lies between $-45^{\circ}$ and $+45^{\circ}$, is 5,786 images (3,816 male images and 1,970 female images). We evaluated our approach using 5 folds of the FERET dataset (700 images were randomly  picked for each fold).

\noindent\textbf{The Specs on Faces (SoF) dataset.} Since one of the main problems in gender classification is the face occlusions and illumination changes \cite{gender,rai2014}, we present a new dataset, the Specs on Faces (SoF)\footnote{\url{http://bit.ly/sof\_dataset}}, that is devoted to these two problems. We made the proposed dataset more challenging for face detection, recognition, and classification, through capturing the faces under harsh illumination environments and face occlusions. The SoF comprises 2,662 original images of size 640 $\times$ 480 pixels for 112 persons (66 males and 46 females) from different ages.  The glasses are the common natural occlusion in all images of the dataset. However, we added two more synthetic occlusions, for nose and mouth, to each image. 

\begin{figure*}[t!]
\centering
\includegraphics[width=1\linewidth]{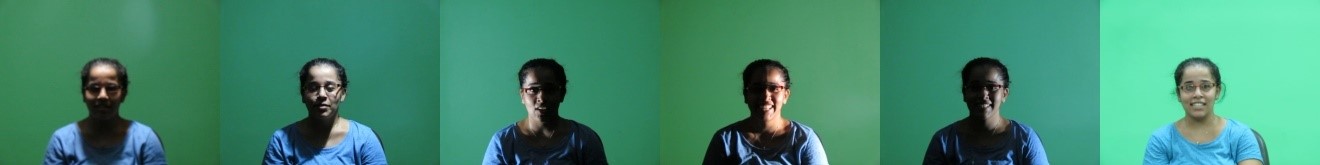}
\caption{\label{fig:sof-part2-sample}Samples of an image for the same person from the Specs on Faces (SoF) data set captured under different lighting directions.}
\end{figure*}

\begin{figure}[t]
\centering
\includegraphics[width=1\linewidth]{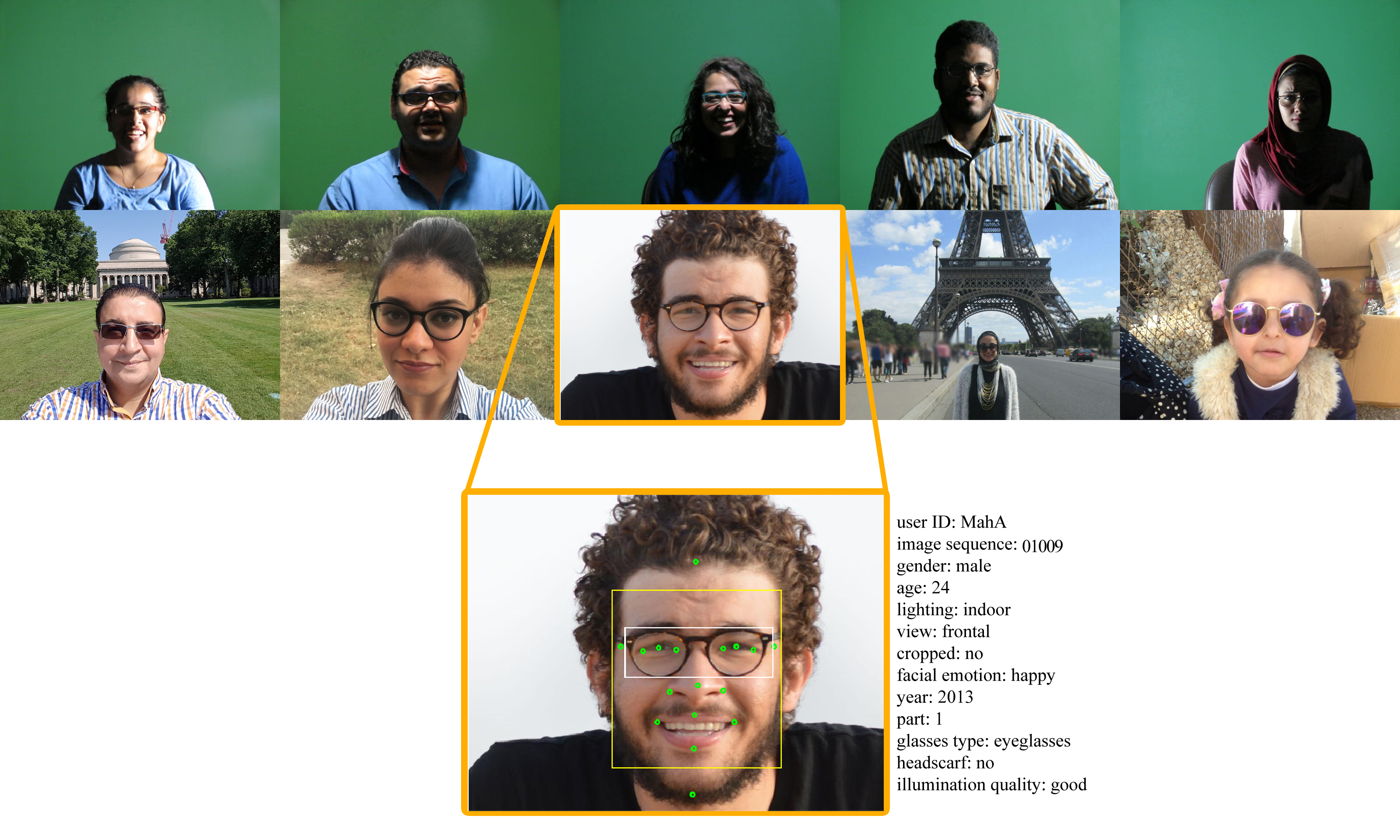}
\caption{\label{fig:sof-metadata}Samples of the Specs on Faces (SoF) dataset. The lower part shows a metadata example for the shown image. The green circles represent the 17 facial landmarks, the white rectangle is the glasses rectangle, and the yellow one is the face rectangle.}
\end{figure}

The original images in the proposed SoF dataset are divided into two parts. The first part contains 757 unconstrained face images in the wild for 106 different persons whose head orientations approximately fall in the range of $\pm 35^{\circ}$ in yaw, pitch, and roll. The images were captured in an unstructured manner over a long period in several locations under indoor and outdoor illumination environments. The second part contains 1905 images which are dedicated to challenging harsh illumination changes. In order to get arbitrary indoor lighting conditions, 12 subjects were captured using a wheel-whirled lamp as the only light source in the laboratory. The lamp is located above and spun around each subject to emit light rays in random directions, see Figure \ref{fig:sof-part2-sample} for an example. This idea is inspired by the primitive version of the Light Stage system presented by Debevec et al. \cite{lightstage}.  The SoF dataset involves a handcrafted metadata that contains subject ID, view (frontal/near-frontal) label, 17 facial feature points, face and glasses rectangle, gender and age labels, illumination quality, and facial emotion for each subject, see Figure \ref{fig:sof-metadata} for an example of the metadata. % The facial expressions were categorized into 4 basic emotions.

\begin{figure}[t]
\centering
\includegraphics[width=.7\linewidth]{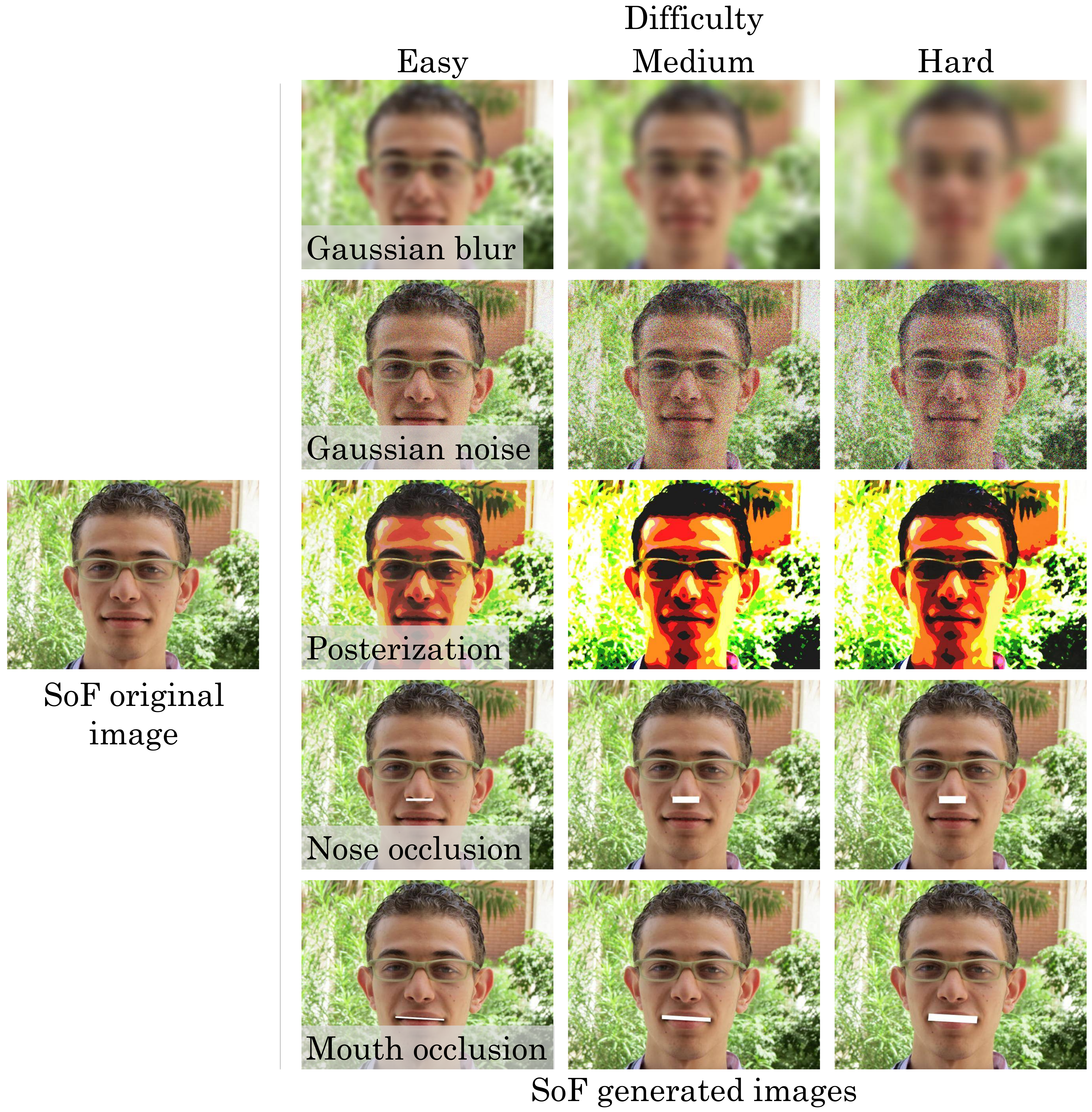}
\caption{\label{fig:sof-sample}The Specs on Faces (SoF) dataset contains two groups of images. The first part includes the original images. The second part contains the original images besides the generated images. The last three columns show the three levels of difficulty (easy, medium, and hard). The rows from the first to the fifth represent the generated images by applying Gaussian smoothing, Gaussian noise, posterization filter, nose occlusion, and mouth occlusion, respectively.}
\end{figure}

Moreover, to generate more challenging synthetic images, we applied three image filters (Gaussian noise, Gaussian smoothing, and image posterization using Fuzzy logic) to the original images. All the generated images are categorized into three levels of difficulty (easy, medium, and hard). That enlarges the number of images to be 42,592 images (26,112 male images and 16,480 female images). Furthermore, the dataset comes with a metadata that describes each subject from different aspects.      Figure \ref{fig:sof-sample} shows a sample image from the two parts of the dataset, original images and synthetic images.

We carried out two groups of experiments using the SoF dataset. In the first group, we randomly picked 5 folds, each contained 330 original images, i.e., without any filters or synthetic occlusions. In the second group, we randomly picked the folds from the whole images of the dataset, original and synthetic images, where each fold contained 750 images.    In the following, we will briefly discuss the results of our facial feature detection mechanism followed by a more thorough discussion of gender classification results achieved by our method.

%We will discuss the details of our proposed dataset in Section \ref{sec:sof}.

%The face images contain various facial expressions for people who wear diverse spectacle frames under different lighting conditions (indoor/outdoor).

\subsection{Facial feature detection}
In spite of the non-frontal view of many images in the datasets, the facial components are extracted in a desirable way. As our target is to extract the facial patch instead of extracting the exact facial points, some error in the alignment of the CDSM is tolerable. In addition, SSR helps improve the feature detection by catching undetected faces, this is shown in Table \ref{Table1}.    As the Face Detection Data Set and Benchmark (FDDB) \cite{FDDB} is devoted to face detection research, it was used to calculate the recall, precision, and F-Measure of the CDSM with and without SSR. The FDDB contains 2845 images that captured 5171 faces. As shown in Table \ref{Table1}, the F-Measure is improved by about $+2\%$ for the FDDB dataset after using SSR as an optional preprocessing step. Also, the SSR improves the F-Measure using SoF (original) and SoF (full) datasets by about $+3\%$ and $+12\%$, respectively.
%The supported annotations of the FDDB dataset are only for the face regions, so the evaluation of the facial feature extraction was performed manually by two persons. 
%The face detection rates of the CDSM/SSR are 99.86\%, 98.11\%, 99.76\%, 99.62\%, 94.67\%, and 81.28\% for the LFW, Groups, Adience, FERET ($\pm 45^{\circ}$), original images of the SoF, and all images of the SoF datasets, respectively.

\begin{table*}[t]
\centering
\caption{The recall (\%), precision (\%), and F-Measure (\%) of the face detection process using cascaded deformable shape model (CDSM) with and without single scale retinex (SSR).}
\label{Table1}
\newcommand{\acm}{1.1cm}
\small
\begin{tabular}{ | L{2cm} | C{1cm} | C{1.3cm} | C{1.2cm} | C{1cm} | C{1.3cm} | C{1.2cm} |}
\hline
 \multirow{3}{*}{Dataset}        & \multicolumn{3}{c|}{CDSM w/o SSR}   & \multicolumn{3}{c|}{CDSM w SSR}     \\ \cline{2-7} 
                 & Recall & Precision & F-Measure 			& Recall 		& 	Precision & F-Measure 		\\ \hline\hline
FDDB            & 73.40  & 99.15    &  84.35        		& 77.99 		& 	97.07 & 86.49   		\\ \hline
SoF (original) 	& 84.08 & 95.30    &  89.34             	&  92.40      	&   93.00 & 92.70      	\\ \hline
SoF (full)    	& 58.86 & 95.24    &  72.76             	& 79.66       	&   90.21 & 84.61 	\\ \hline
\end{tabular}
\end{table*} 

\subsection{Gender classification accuracy}
We have applied our proposed method (AFIF$^4$) to unconstrained types of face images, i.e., frontal images, near-frontal images, non-frontal images, and images with large poses and occlusions. In literature, many gender classification methods are applied only to frontal or near-frontal face images \cite{moeini2015real, mery2015automatic, tapia2013, shan2012learning, baluja2007boosting}.    For the sake of fair comparison, we report only results of methods using unconstrained types of images \cite{moeini2017, hadid2015, levi2015cnn, van2015deep, han2015demographic, rai2014, Adience, groups} and we omit results of methods using only frontal or near-frontal face images. From the work by Moeini and Mozaffari \cite{moeini2017}, we report the results of two methods: dictionary learning and separate dictionary learning for gender classification, denoted as DL-GC and SDL-GC, respectively.  Only for the case of the FERET dataset, because we used frontal and near-frontal images with pose angles between $-45^{\circ}$ and $+45^{\circ}$,  we report results of methods using frontal/near-frontal images.    Table \ref{Table2} shows that the accuracy of our method outperforms the state-of-the-art results reported for unconstrained types of face images from the LFW, Groups, and Adience datasets. Also, our method achieves comparable accuracy with the state-of-the-art over the FERET dataset for frontal/near-frontal face images.

%Methods using only frontal and near-frontal face images\\
%2015: \cite{moeini2015real}, \cite{mery2015automatic}: 93.3 groups, \\
%2013: \cite{tapia2013}\\
%2012: \cite{shan2012learning}\\
%2007: \cite{baluja2007boosting}\\
%
%Methods using images with small poses:\\
%2015: \cite{han2015demographic}\\
%
%Methods using unconstrained face images:\\
%2017: \cite{moeini2017}\\
%2015: \cite{hadid2015}, \cite{levi2015cnn}, \cite{van2015deep}\\
%2014: \cite{rai2014}, \cite{Adience}\\
%2009: \cite{groups}\\

\begin{table*}[t]
\centering
\caption{Comparison of our method (AFIF$^4$) with state-of-the-art achieved accuracy over the LFW, Groups, Adience, and FERET datasets. Note that the reported accuracy here are for methods applied to unconstrained types of images (frontal, near-frontal, and images with large poses and occlusions), results for only frontal and/or near-frontal images are omitted for the sake of fair comparison.  The cells marked with a `--' represent unavailable results or results from methods using only frontal/near-frontal images. Only for the case of FERET dataset, we report results of methods using only frontal/near-frontal images because we followed the same procedure.}
\label{Table2}
\small
\newcommand{\wa}{1.5cm}
\begin{tabular}{| L{3.5cm} | C{\wa} | C{\wa} | C{\wa} | C{\wa} |}
\hline
\multirow{2}{*}{Method}             & \multicolumn{4}{c|}{Accuracy (\%)} \\ \cline{2-5}
			                        & LFW            & Groups         & Adience       &  FERET \\ \hline\hline
DL-GC \cite{moeini2017}             & 93.60          & 84.40          & --     		  & 99.50  \\ \hline
SDL-GC \cite{moeini2017}            & 94.90          & 83.30   		  & --     		  & \gbc{99.90}  \\ \hline
Hadid et al. \cite{hadid2015}       &   --           & 89.85          & --            & --     \\ \hline
Eidinger et al. \cite{Adience}      &   --           & 86.80          & 76.10         & --     \\ \hline
Han et al. \cite{han2015demographic}& 94.40          &   --           & --            & --     \\ \hline
Rai and Khanna \cite{rai2014}		& 89.10          &   --           & --            & 98.40  \\ \hline
Gallagher and Chen \cite{groups}    &  --            & 74.10          & --            & --     \\ \hline
Levi and Hassner \cite{levi2015cnn} & --             & --             & 86.80         & --     \\ \hline
Wolfshaar et al. \cite{van2015deep} & --             & --             & 87.20         & --     \\ \hline
Tapia and P\'erez \cite{tapia2013}    & --             & --             & --            & 99.10     \\ \hline
AFIF$^4$ (Ours)  	                & \gbc{95.98} & \gbc{90.73} & \gbc{90.59} & 99.49 \\ \hline
\end{tabular}
\end{table*}

% # of test images
%         LFW  Groups  Adience  FERET  SoF (original)   SoF (full)
% AFIF:   590  2536    970      700    330              750   

\noindent\textbf{Cross-dataset Evaluation.} To further assess the performance of our method, we carried out \textit{cross-dataset} evaluation as shown in Table \ref{Table3}.   The attained accuracy using the same dataset for both training and testing usually drops in an obvious way when using different datasets for training and testing. From Table \ref{Table3}, we can see that the lowest cross-dataset classification accuracy is obtained using the full SoF dataset (in 3 cases), the Adience dataset (in 2 cases), and the Groups dataset (in 1 case). This points towards that our full SoF dataset is the most challenging. The low accuracy obtained using the full SoF dataset, compared to the other datasets, is due to the challenging filters and synthetic occlusions that have been added to the original images. Also, it is worth noting that the highest cross-dataset accuracy is obtained by the FERET dataset, due to the good quality of the images compared with the poor resolutions of many images of other datasets. 

\begin{table*}[]
\centering\small
\caption{Results of cross-dataset evaluation of our proposed method (AFIF$^4$). The diagonal represents the average accuracy obtained using the same dataset for both training and testing. Rows represent datasets used for training while columns represent datasets used for testing. The values in bold represents the dataset that yields the lowest accuracy when using a specific dataset for training, for example, in first row, when using LFW for training, the full SoF dataset yields the lowest accuracy, that means it is the most challenging in this case.}
\label{Table3}
\newcommand{\bw}{1.1cm}
\begin{tabular}{| L{2cm} | C{\bw} | C{1.3cm} | C{\bw} | C{1.5cm} | C{\bw} | C{\bw} |}
\hline
  & \multicolumn{6}{c|}{\textbf{Testing Accuracy (\%)}} \\ \cline{2-7} 
 \textbf{Training}            & LFW       &   Adience  &  SoF (full)  &   SoF (original)   &    Groups    &  FERET \\ \hline\hline
 LFW   & \cellcolor{gray!25}{95.98} & 79.19  & \gbc{65.76}   & 78.36    & 76.67      & 92.71            \\ \hline
 Adience          & 84.55 & \cellcolor{gray!25}{90.59}  & \gbc{74.15}   & 79.97    & 85.07   & 86.86   \\ \hline
 SoF (full)       & 79.22 & 74.16  & \cellcolor{gray!25}{92.10}   & 97.21    & \gbc{72.30}     & 84.77     \\ \hline 
SoF (original)    & 83.31 & \gbc{71.09}  & 84.05   & \cellcolor{gray!25}{98.48}    & 73.60     & 89.12      \\ \hline
Groups            & 91.74 & 83.06  & \gbc{69.80}   & 82.20   & \cellcolor{gray!25}{90.73}    & 92.20      \\ \hline 
FERET             & 85.78 & \gbc{69.19}   & 75.15   & 83.27    & 85.67     & \cellcolor{gray!25}{99.49}      \\ \hline
\end{tabular}
\end{table*}

To assess the performance for cross-dataset evaluation against the state-of-the-art performance, we compare our results with the latest results reported by Moeini and Mozaffari \cite{moeini2017} for the LFW, Groups, and FERET datasets in Table \ref{Table4}. It is clear that our method gives higher accuracy for all cases except for the FERET dataset. It is worth noting that cross-dataset evaluation usually yields lower accuracy than the case of using the same dataset for both training and testing. This is mainly due to different conditions of collecting images in different datasets, such as occlusions, illumination changes, backgrounds, etc.

\begin{table*}[]
\centering\small
\caption{Comparison of cross-dataset evaluation of our proposed method (AFIF$^4$) against the state-of-the-art results reported by Moeini and Mozaffari \cite{moeini2017}. The values represent the classification accuracy (\%).}
\label{Table4}
\newcommand{\bw}{1.5cm}
\begin{tabular}{| L{\bw} | C{\bw} | C{\bw} | C{\bw} | C{\bw} |}
\hline 
 Training & Testing & DL-GC \cite{moeini2017} & SDL-GC \cite{moeini2017}  &  AFIF$^4$ (Ours)\\\hline\hline
LFW & LFW & 93.60 & 94.90 & \gbc{95.98}\\ \hhline{|~|*{4}{-|}}
    & Groups & 69.70 & 72.40 & \gbc{76.67} \\ \hhline{|~|*{4}{-|}}
    & FERET & 88.80 & 89.70 & \gbc{92.71}\\ \hline
Groups & LFW & 80.60 & 83.10 & \gbc{91.74}\\ \hhline{|~|*{4}{-|}}
    & Groups & 83.30 & 84.40 & \gbc{90.73} \\ \hhline{|~|*{4}{-|}}
    & FERET & 85.20 & 87.10 & \gbc{92.20}\\ \hline
FERET & LFW & 71.70 & 73.50 & \gbc{85.78}\\ \hhline{|~|*{4}{-|}}
    & Groups & 59.50 & 61.90 & \gbc{85.67} \\ \hhline{|~|*{4}{-|}}
    & FERET & 99.50 & \gbc{99.90} & 99.49\\ \hline
\end{tabular}
\end{table*}

%%%%%%%%%%%%%%%%%%%%%%%%%%%%%%%%%%%%%%%%%%%%%%%%%%%%%%%%%%%%%%%%

\section{Conclusion}\label{sec:conclusion}

In this paper, we addressed the gender classification problem by using a combination between local and holistic features extracted from face images. We used four deep convolutional neural networks (CNNs) to separately classify the individual features, then we applied an AdaBoost-based score fusion mechanism to aggregate the prediction scores obtained from the CNNs. Through extensive experiments, we showed that our method achieves better results than the state-of-the-art methods in most cases on widely-used datasets. Also, and more importantly, we showed that our method performs better than the state-of-the-art when generalized to cross-dataset evaluation, which is much more challenging than in-dataset evaluation.             Furthermore, we proposed a more challenging dataset of 42,592 face images that mainly addresses the challenges of face occlusions and illumination variation.  We accompanied our proposed dataset with handcrafted annotations and gender labels for all images to facilitate further research addressing the gender classification problem.

%\paragraph{Installation} 
%\emph{elsarticle} 

\end{document}